# USU-Corn-WeedDB: A UAV RGB Image Dataset for Multi-Species Weed Detection in Forage Corn

*Utsav Bhandari[1], Saroj Burlakoti[2], Rhonda Miller[1], Sierra Young[3], Eric Westra[2], Aaron Etienne[1]*

*[1]Department of Applied Sciences, Technology, and Education, Utah State University, Logan, UT, USA*
*[2]Department of Plants, Soils & Climate, Utah State University, Logan, UT, USA*
*[3]Department of Civil and Environmental Engineering, Utah State University, Logan, UT, USA*

**Abstract:**

Weed pressure in forage corn production causes yield losses of up to 31.5%, yet site-specific weed management (SSWM) systems built on UAV imagery and deep learning remain constrained by the scarcity of field-representative training datasets. We present USU-Corn-WeedDB, a publicly available UAV RGB image dataset collected from a commercial forage corn field in Cache Valley, Utah, designed to support multi-class weed detection under both supervised and semi-supervised learning frameworks. RGB imagery was acquired on 27 June 2025 using an Autel EVO II Dual 640T V2 drone at ~10m above ground level, yielding a ground sampling distance of approximately 0.48 cm/pixel. A total of 366 full-resolution images were tiled into 8,800 patches at 640 × 640-pixel resolution. Of these, 800 images were manually annotated for three weed species; common lambsquarters *(Chenopodium album)*, redroot pigweed *(Amaranthus retroflexus)*, and green foxtail *(Setaria viridis)* comprising 10,539 bounding-box instances, with the remaining 8,000 tiles retained as an unlabeled pool for semi-supervised experiments. This dataset reflects a natural class imbalance where redroot pigweed constitutes 53.86% of annotated instances, which was preserved intentionally to mirror real field conditions. To validate dataset utility, we trained 28 object detection models spanning five architecture families including YOLOv8, YOLOv9, YOLOv10, YOLO11, YOLO26, and RT-DETR under identical conditions without hyperparameter tuning. Test set mAP@0.5 ranged from 0.773 to 0.840, with lightweight models achieving competitive performance relevant to edge-deployed UAV systems. USU-Corn-WeedDB is publicly available at https://doi.org/10.5281/zenodo.20044178.

**List of abbreviations: CL**: Common Lambsquarters, **CVAT**: Computer Vision Annotation Tool, **GFT**: Green Foxtail, **mAP**: Mean Average Precision, **RGB**: Red, Green, and Blue, **RPW**: Redroot Pigweed, **SSWM**: Site-Specific Weed Management, **SSL**: Semi-Supervised Learning, **UAV**: Unmanned Aerial Vehicle, **YOLO**: You Only Look Once

## 1. Background

Weed pressure remains one of the most significant challenges in crop production, with annual yield losses up to 31.5% [1]. Site-specific weed management (SSWM) systems built on unmanned aerial vehicle (UAV) imagery and deep learning-based detection offer a pathway to weed management with efficient use of herbicides [2]. Yet despite meaningful progress in UAV based detection systems over the past several years, deployment in real production fields consistently exposes a fundamental bottleneck: model performance is bounded not by the algorithm, but by the quality and representativeness of the training data [3]. A detection model trained under one set of field conditions often fails under another, as RGB imagery is sensitive to variability in soil color, weed phenotype, canopy structure, and illumination [4]. These variations can be partially mitigated through data augmentation techniques; however, such methods still cannot fully replicate the true variability observed at the field level [5]. These constraints suggest that dataset construction is not just a preprocessing step, but a core component that determines the success of any detection system trained on it.

Field-specific dataset creation remains a major bottleneck in agricultural deep learning, demanding significant time and expert labor for accurate data annotation. This is especially prevalent in corn fields. Weeds such as common lambsquarters emerge early and aggressively alongside the crop. In contrast, species like redroot pigweed and green foxtail tend to emerge progressively after crop establishment. Together, this temporal variability in emergence creates an annotation challenge that demands continuous expert attention [1]. Annotators must distinguish morphologically similar species with overlapping canopy structures, navigate partially occluded and densely clustered weeds, and maintain consistent labeling standards across thousands of image tiles captured under varying flight altitudes, illumination, and growth stages. Thus, a single field may require months of expert labor to achieve accurate species-level annotations [4,5]. Collectively, these constraints suggest that manual annotation represents one of the most limiting factors in large-scale agricultural dataset preparation.

Semi-supervised learning (SSL) offers a practical solution to the challenge of manual annotation [3,6]. In this approach, an initial model is trained on a small, carefully annotated subset of data. It then iteratively generates pseudo-labels for unannotated images using various SSL frameworks. This process can substantially reduce the annotation burden while still retaining the benefits of label-guided supervision. The effectiveness of this approach, however, depends on the availability of a paired dataset: a labeled partition large enough to yield a credible initial model, and an unlabeled partition large enough to be worth pseudo-labeling [7,8]. Such paired resources are rarely made publicly available in the weed detection literature, which limits the community's ability to benchmark SSL approaches under commercial field conditions.

This paper addresses both challenges. We present USU-Corn-WeedDB, a publicly available dataset comprised of annotated and unannotated UAV imagery collected from commercial forage corn fields in Utah. This dataset is designed to support multi-class weed detection model development under both fully supervised and semi-supervised training approaches. It provides researchers with a resource for studying generalization, annotation efficiency, and SSL model performance in real-world commercial field conditions.

## 2. Methodology

### *2.1. Experimental Site*

The data used in this study was collected at Utah State University's Cache Junction research farm in Cache Valley, Utah, USA (41.82 °N, 111.98 °W, elevation 1,349 m). The specific field used was a commercial forage corn plot located in the semi-arid Intermountain West region. The field soil is classified as the Trenton silty clay loam series (USDA map unit TrA), formed from lacustrine deposits derived from quartzite, limestone, and sandstone on a nearly level lake terrace (0-2% slope). The profile is dominated by silty clay from 20 cm depth downward, underlain by a seasonal water table at roughly 107-152 cm. Drainage is somewhat poor, with very low to moderately low saturated hydraulic conductivity, placing this soil in USDA Hydrologic Soil Group D with a land capability class of 4w under irrigation [9]. During the 2025 growing season (May-October), the site received approximately 102 mm of natural precipitation and was delivered a total of 433.6 mm of supplemental irrigation through a center-pivot irrigation system. The average daily temperatures ranged from 11.3 °C in October to 23.1 °C in July, with July daily maximum temperatures averaging 33.2 °C. Wind was predominantly out of the southwest at an average of 1.82 m/s. The forage corn on the specific field was planted on 23 May 2025.

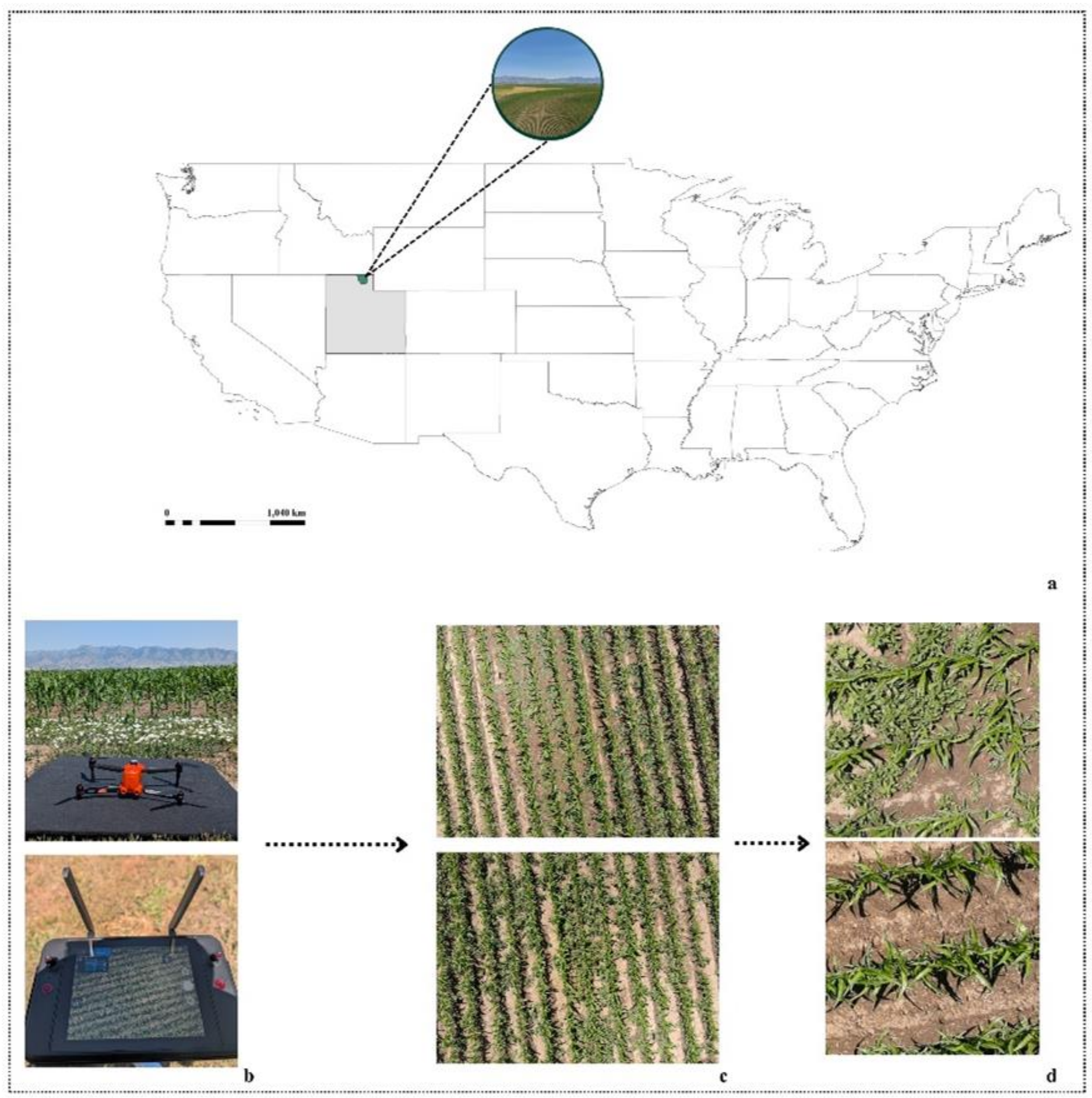

*Figure 1: (a) Location of the field (b) Autel Drone used for image acquisition (c) RGB Image acquired using the drone (d) Tiled images for annotations*

### *2.2. Dataset Acquisition*

Images were captured using an Autel EVO II Dual 640T V2 unmanned aerial vehicle equipped with a 48-megapixel RGB camera (1/2-inch CMOS sensor) [10]. Weekly flights were conducted throughout June 2025 to document plant and weed development across multiple growth stages. Following inspection of all collected imagery, the flight conducted on 27 June 2025 was selected for dataset development, as it captured the most consistent weed emergence and canopy differentiation relative to the corn crop. Data collection on that date took place between 14:43 and 14:57 local time (Mountain Standard Time [MST]) under clear conditions: air temperature 31.39 °C, relative humidity 15.5%, wind speed 4.36 m/s, and solar radiation 906 W/m$^2$. No precipitation was recorded during or immediately before the flight. The flight was conducted at ~10m above ground level at an air speed of approximately 3 m/s, yielding a ground sampling distance of approximately 0.32 cm/pixel. At the time of acquisition, the corn crop was at the V4 growth stage, and weed pressure was visible throughout the field. A total of 324 RGB images were captured at 4000 × 3000-pixel resolution. These full-resolution images were tiled into smaller patches of 640 × 640-pixel resolution using a custom Python script, blurred images were removed, producing a final dataset of 8,800 tiles for model development. This tiling of high-resolution UAV imagery is a standard preprocessing step in remote sensing and computer vision, as it preserves native spatial resolution rather than downscaling the imagery [8]. It has also proven to be faster for developing the models, and it increases the number of training samples available to the model [8]. Each tile retained the full detail of the original capture, which is important for detecting small weed objects that would otherwise lose definition at reduced input resolutions.

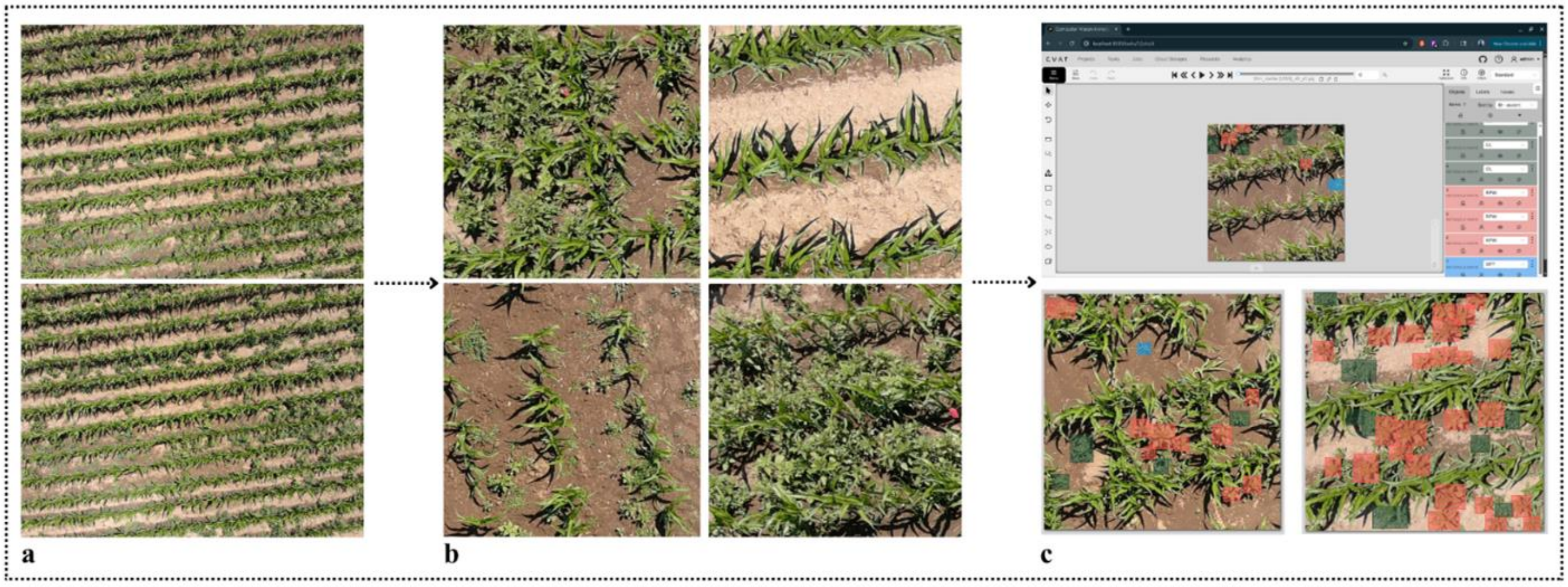

*Figure 2: (a) Original 4000 × 3000-pixel resolution image taken with drone (b) 640 × 640-pixel resolution tiled images (c) Annotations in CVAT (CVAT-UI: up, annotations (down) (Grey: common lambsquarters, Red: redroot pigweed, Blue: green foxtail)*

***2.3. Dataset Preparation***

The tiled images were uploaded to the Computer Vision Annotation Tool (CVAT), an open-source web-based annotation platform widely used in computer-vision and remote-sensing research (https://www.cvat.ai/). A total of 650 images were manually annotated for three weed species: common lambsquarters (*Chenopodium album*), redroot pigweed (*Amaranthus retroflexus*), and green foxtail (*Setaria viridis*). Annotations were drawn as tight bounding boxes and reviewed in consultation with weed science faculty and graduate students at Utah State University to minimize labeling errors prior to model development.

To reduce false positives and help the model distinguish between crop and weed classes, 150 additional background images containing only corn plants with no target weeds present were included as hard negative samples, a common practice in single-stage object-detection workflows where target classes share visual space with a dominant non-target class [11]. The annotated dataset was exported in the Ultralytics YOLO detection format, with bounding-box coordinates stored as `.txt` files paired with each image. The complete dataset of 800 images and their corresponding labels was split into training, validation, and test subsets for model development. The same split was used for all supervised and semi-supervised experiments to ensure fair comparison. The training set was used to fit model weights; the validation set was used for hyperparameter tuning and early-stopping decisions; and the test set was held out entirely for final evaluation and was not used during training or validation. The remaining 8,000 tiled images were retained as an unlabeled dataset for use in the semi-supervised object detection experiments. The full dataset is available in https://doi.org/10.5281/zenodo.20044178 and its composition is summarized in Table 1.

The dataset shows a severe class imbalance, with redroot pigweed (RPW) representing 53.86% of all annotated instances; more than double the proportion of common lambsquarters (22.14%) and green foxtail (24.01%). No corrections, such as oversampling, undersampling, or class-weighting were applied to address this imbalance. In commercial production fields, weed populations are rarely evenly distributed, and one species often dominates depending on local soil and field conditions. Artificially balancing the class distribution would not reflect what farmers actually encounter in their fields. The natural species distribution was therefore retained to ensure that the dataset accurately represents real-world weed pressure in Intermountain West forage corn production.

*Table 1:Complete Dataset Information*

| Metric | Value |
|---|---|
| **Class distribution** | |
| CL: common lambsquarters (class 0) | 2,333 instances |
| RPW: redroot pigweed (class 1) | 5,676 instances |
| GFT: green foxtail (class 2) | 2,530 instances |
| **Annotated dataset statistics** | |
| Total labeled dataset size | 800 |
| Total annotations | 10,539 |
| Background images (hard negatives) | 150 |
| Avg. annotations per image | 13.17 |
| Annotation format | YOLO (.txt) |
| **Dataset split** | |
| Training | 660 images (8,490 instances: 80.6%) |
| Validation | 70 images (1,103 instances: 10.5%) |
| Test | 70 images (946 instances: 8.9%) |
| **Unlabeled Images** | |
| Total unlabeled images | 8000 images |

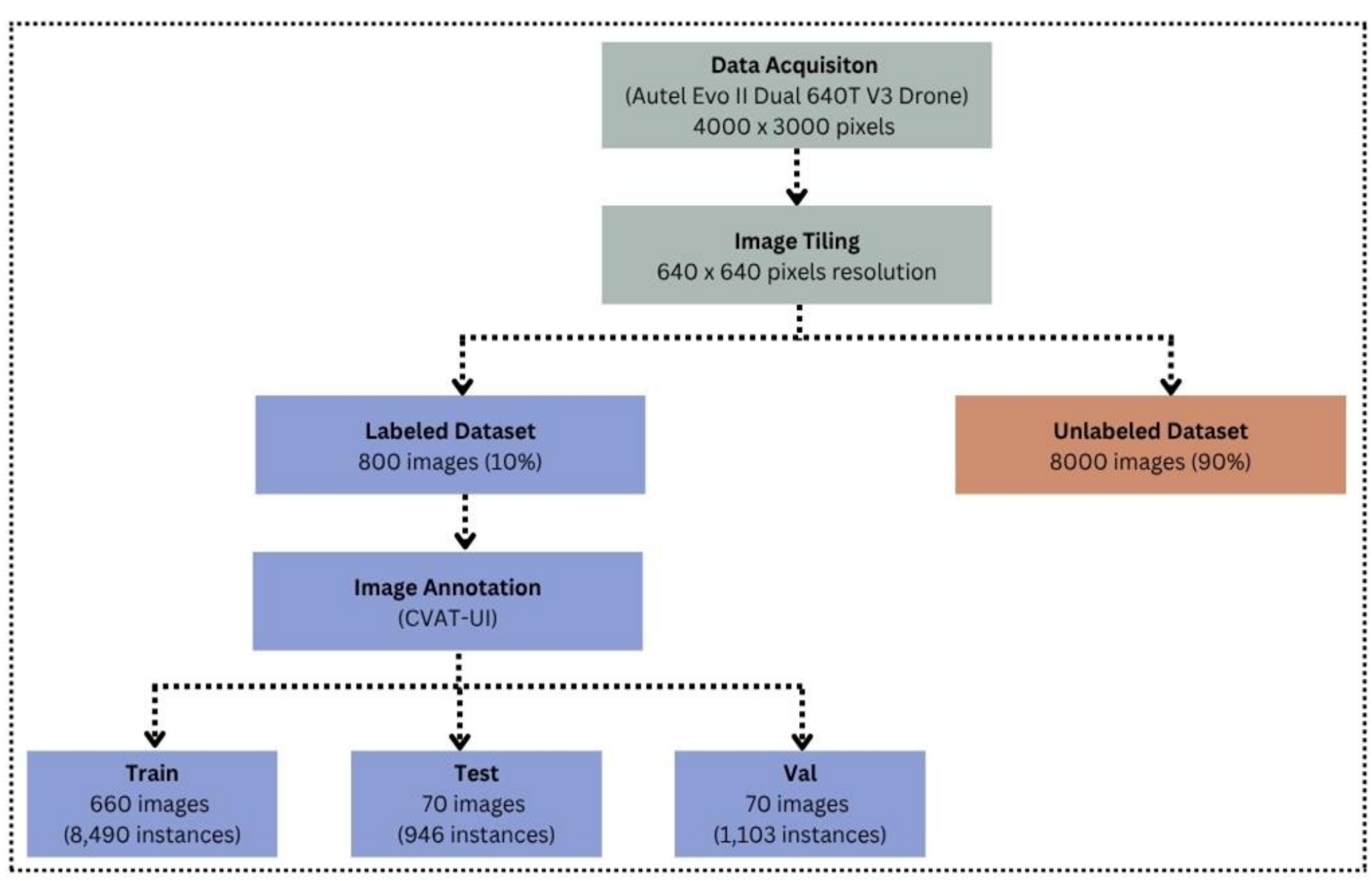

*Figure 3: Overall dataset preparation process and structure*

### 3. Technical validation

To validate the utility of USU-Corn-WeedDB for supervised weed detection, a benchmark evaluation was conducted using 28 single-stage object detection models spanning five architecture families: YOLOv8, YOLOv9, YOLOv10, YOLO11, YOLO26, and RT-DETR. All models were trained on the designated training split and evaluated on both the validation and held-out test sets using standard detection metrics: precision, recall, F1-score, mAP@50, and mAP@50-95. No hyperparameter tuning was applied beyond default settings, ensuring the results reflect the dataset's baseline difficulty rather than model-specific optimization.

On the test set, mAP@50 scores ranged from 0.773 (YOLO26n) to 0.840 (YOLOv9s), indicating consistent detectability of weed instances across model scales. The mAP@50-95 range of 0.423-0.480 reflects the inherent challenge of precise localization under natural field conditions, including overlapping canopies and variable weed sizes. Notably, lightweight models such as YOLOv9t (4.43 MB) and YOLOv9s (14.53 MB) achieved competitive mAP@50 scores of 0.832 and 0.840 respectively, demonstrating that the dataset supports effective learning even without large model capacity; an important consideration for UAV-based edge deployment. RT-DETR variants performed comparably to YOLO-family models, with RT-DETR-x achieving a test mAP@50 of 0.814 and the highest precision among all models (0.814). These results confirm that USU-Corn-WeedDB can effectively support the training of weed detection models, and the reported metrics provide a baseline reference for future research using this dataset [12].

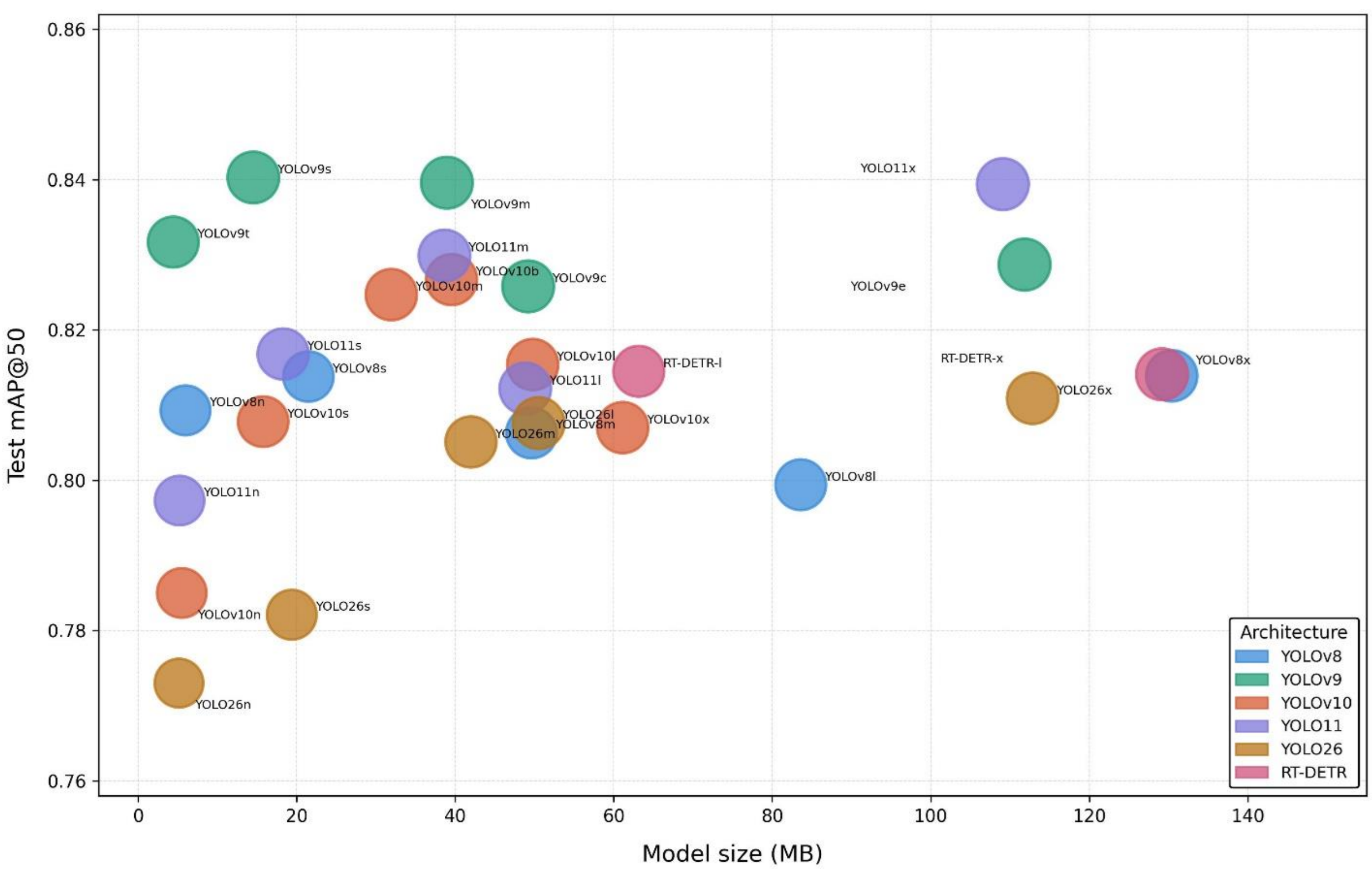

*Figure 4: Bubble chart showcasing the test set mAP@50 and model size in MB across all 28 trained models*

**4. Data availability**
The full dataset is publicly available in: https://doi.org/10.5281/zenodo.20044178

**5. Limitations**
Several limitations should be considered when utilizing this dataset. All images were collected from a single field in Cache Valley, Utah, on a single acquisition day at a fixed altitude of ~13 m above ground level under clear midday conditions. This means the dataset captures neither geographic nor temporal variability across crop and weed growth stages; thus, trained model performance may degrade under overcast illumination, different flight altitudes, or alternative viewing angles. The natural class imbalance, with redroot pigweed comprising 53.86% of annotated instances, mirrors real field conditions but may cause models to underperform on minority classes like common lambsquarters and green foxtail.

**CRediT Author Statement**

**Utsav Bhandari:** Conceptualization, Methodology, Visualization, Data collection, Data annotation, Writing-Original draft preparation, Review and Editing. **Saroj Burlakoti:** Methodology, Data annotation, Writing- Review & Editing, **Rhonda Miller:** Supervision **Sierra Young:** Supervision, Resources. **Eric Westra:** Supervision, Conceptualization, Resources, Methodology. **Aaron Etienne:** Conceptualization, Resources, Data collection, Supervision, Project Administration, Funding acquisition, Writing- Reviewing & Editing.

**Funding**

Funding was made available for this study from the Utah Agricultural Experiment Station Project 1832.